\documentclass{article}

\usepackage{arxiv}

\usepackage[utf8]{inputenc} 
\usepackage[T1]{fontenc}    
\usepackage{hyperref}       
\usepackage{url}            
\usepackage{booktabs}       
\usepackage{amsfonts}       
\usepackage{nicefrac}       
\usepackage{microtype}      
\usepackage{lipsum}		
\usepackage{graphicx}
\usepackage{natbib}
\usepackage{doi}
\usepackage{multirow}

\title{ANNETTE: Accurate Neural Network Execution Time Estimation with Stacked Models}

\author{ \href{https://orcid.org/0000-0002-1877-4114}{\includegraphics[scale=0.06]{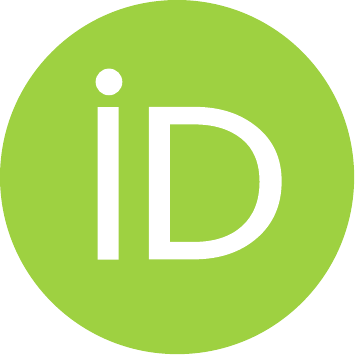}\hspace{1mm}Matthias Wess} \\
	TU Wien, Institute for Computer Technology \\
	\texttt{matthias.wess@tuwien.ac.at} \\
	\And
	Matvey Ivanov \\
	TU Wien, Institute for Computer Technology \\
	\And
	Anvesh~Nookala \\
	TU Wien, Institute for Computer Technology \\
	\And
	Christoph Unger \\
	TU Wien, Institute of Automation and Control \\
	\And
	Alexander~Wendt \\
	TU Wien, Institute for Computer Technology \\
	\And
	Axel Jantsch \\
	TU Wien, Institute for Computer Technology \\
}

\hypersetup{
pdftitle={Accurate Neural Network Execution Time Estimation with Stacked Models},
pdfsubject={cs.AR},
pdfauthor={Matthias Wess, Matvey Ivanov, Christoph Unger, Anvesh Nookala, Alexander Wendt, Axel Jatnsch},
pdfkeywords={Analytical models, Estimation, Neural network hardware},
}

\usepackage[acronym]{glossaries} 
\glsdisablehyper 
\newacronym{admm}{ADMM}{Alternating Direction Method of Multipliers}
\newacronym{nn}{NN}{Neural Network}
\newacronym{ai}{AI}{Artificial Intelligence}
\newacronym{cnn}{CNN}{Convolutional Neural Network}
\newacronym{dnn}{DNN}{Deep Neural Network}
\newacronym{lstm}{LSTM}{Long-short Term Memory}
\newacronym{gan}{GAN}{Generative Adversarial Network}
\newacronym{gru}{GRU}{Gated Recurrent Unit}
\newacronym{svd}{SVD}{Singular Value Decomposition}
\newacronym{sgd}{SGD}{Stochastic Gradient Descent}
\newacronym{mil}{MIL}{Multiple Instance Learning}
\newacronym{fpga}{FPGA}{Field Programmable Gate Array}
\newacronym{gpu}{GPU}{Graphic Processing Unit}
\newacronym{cpu}{CPU}{Central Processing Unit}
\newacronym{mgpu}{mGPU}{mobile Graphic Processing Unit}
\newacronym{npu}{NPU}{Neural Processing Unit}
\newacronym{vpu}{VPU}{Vision Processing Unit}
\newacronym{sdk}{SDK}{Software Development Kit}
\newacronym{gemm}{GEMM}{General Matrix Multiply}
\newacronym{tdp}{TDP}{Thermal Design Power}
\newacronym{adas}{ADAS}{Advanced Driver-Assistance Systems}
\newacronym{mac}{MAC}{Multiply-Accumulate operation}
\newacronym{flop}{FLOP}{Floating Point Operation}
\newacronym{nas}{NAS}{Neural Architecture Search}
\newacronym{ncs2}{NCS2}{Intel Neural Compute Stick 2}
\newacronym{dnndk}{DNNDK}{Xilinx Deep Neural Network Development Kit}
\newacronym{dpu}{DPU}{Deep Neural Network Processing Unit}
\newacronym{ops}{OP}{Operation}
\newacronym{height}{h}{height}
\newacronym{width}{w}{width}
\newacronym{channels}{c}{number of input channels}
\newacronym{filters}{f}{number of filters}
\newacronym{annette}{ANNETTE}{Accurate Neural Network Execution Time Estimation}
\newacronym{rmspe}{RMSPE}{Root-Mean-Square-Percentage-Error}
\newacronym{rmse}{RMSE}{Root-Mean-Square-Error}
\newacronym{mape}{MAPE}{Mean-Absolute-Percentage-Error}
\newacronym{mae}{MAE}{Mean-Absolute-Error}
\newacronym{lmt}{LMT}{Linear Model Tree}
\newacronym{mcc}{MCC}{Matthews Correlation Coefficient}
\newacronym{pe}{PE}{Processing Element}

\begin{document}
\maketitle

\begin{abstract}

With new accelerator hardware for \glspl{dnn}, the computing power for \gls{ai} applications has increased rapidly. However, as \gls{dnn} algorithms become more complex and optimized for specific applications, latency requirements remain challenging, and it is critical to find the optimal points in the design space. To decouple the architectural search from the target hardware, we propose a time estimation framework that allows for modeling the inference latency of \glspl{dnn} on hardware accelerators based on mapping and layer-wise estimation models. The proposed methodology extracts a set of models from micro-kernel and multi-layer benchmarks and generates a stacked model for mapping and network execution time estimation. We compare estimation accuracy and fidelity of the generated mixed models, statistical models with the roofline model, and a refined roofline model for evaluation. We test the mixed models on the ZCU102 SoC board with \gls{dnndk} and \gls{ncs2} on a set of 12 state-of-the-art neural networks. It shows an average estimation error of 3.47\% for the \gls{dnndk} and 7.44\% for the \gls{ncs2}, outperforming the statistical and analytical layer models for almost all selected networks. For a randomly selected subset of 34 networks of the NASBench dataset, the mixed model reaches fidelity of 0.988 in Spearman's $\rho$ rank correlation coefficient metric. The code of ANNETTE is publicly available\footnote{https://github.com/embedded-machine-learning/annette}.

\end{abstract}



\section{Introduction}
\label{sec:intro}
Deep Neural Networks have become key components in many \gls{ai} applications, including autonomous driving~\cite{grazer2020}, medical diagnosis ~\cite{Rajpurkar2017,wess2017neural} and machine translation \cite{Zhang2015a}. The computational intensity of some \gls{ai} applications based on \glspl{dnn} prevents their use on embedded system platforms, as these algorithms often have to meet latency and performance requirements to fulfill their purpose.

Attempting to close the gap between the computational intensity of \glspl{dnn} and the available computing power, a wide variety of hardware accelerators for \glspl{dnn} and other \gls{ai} workloads have emerged in recent years. A considerable amount of research has improved the efficiency of \glspl{dnn} and reduced their memory consumption by applying methods such as pruning \cite{tung2018clip}, \cite{srinivas2015data}, quantization~\cite{wess2018weighted,icg:shin2016fixed,icg:miyashita2016log}, and factorization~\cite{icg:jaderberg2014speeding,icg:tai2016lowrank}. Alternatively, a network architecture that is expected to work efficiently on the target device can be designed and trained directly. Networks like MobileNet \cite{icg:sandler2018mobilenetv2} and ShuffleNet \cite{zhang2018shufflenet} are specifically designed to reduce the number of \glspl{mac}, but they contain specific layer types that are not necessarily optimal for all hardware types. In addition, computational efficiency depends largely on the specific architectural parameters of each layer and the hardware platform used~\cite{Yang2018a}. 

\begin{figure}[ht]
  \centering
  \includegraphics[width=0.7\textwidth]{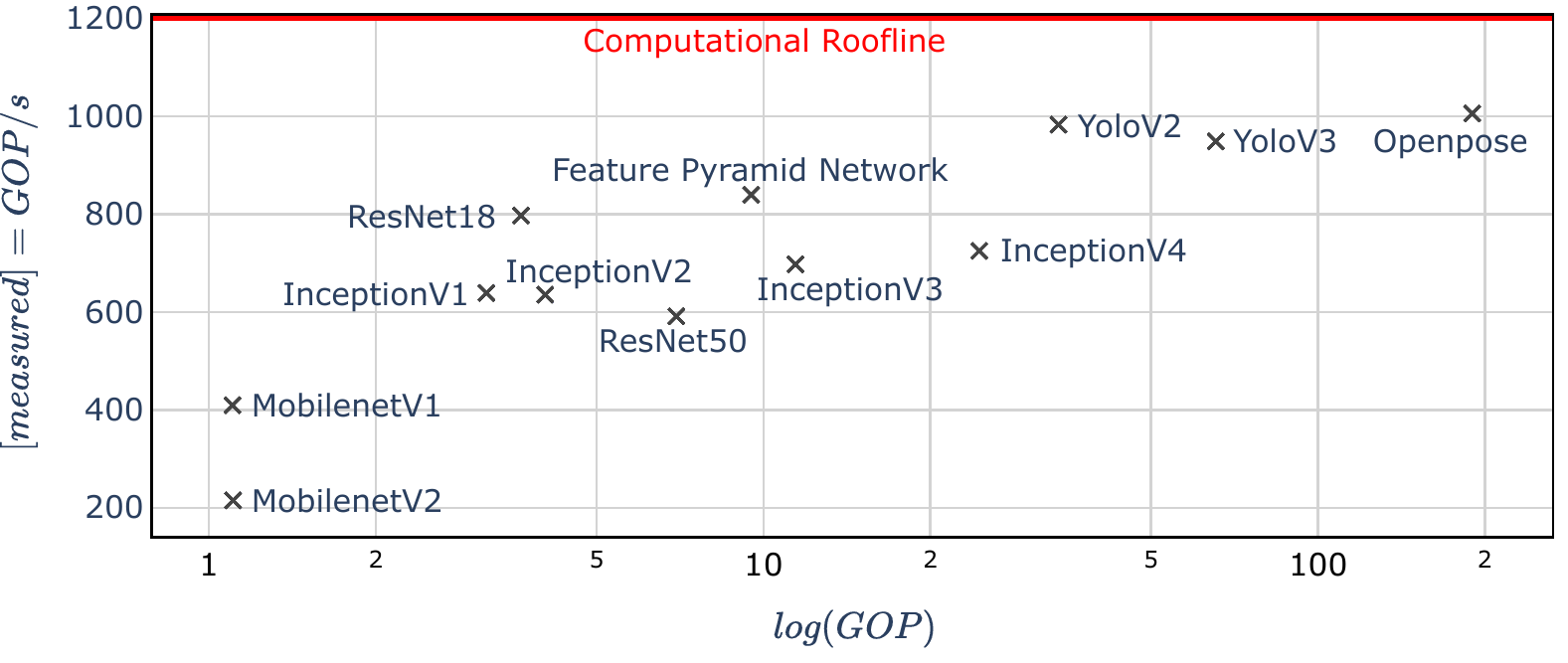}
  \caption{Effective compute performance when inferring the \glspl{dnn} from Table~\ref{tab:networks} on the Xilinx ZCU102 evaluation board. }
  \label{fig:networks}
\end{figure}

Finally, also the mapping toolchain optimizing the original network graph for the selected hardware platform has to be considered since many hardware
accelerators allow specific combinations of layers to be fused together to reduce inter-layer data transfer and/or to optimize data flow. Therefore, when optimizing the network architecture towards "direct metrics" such as latency or energy consumption, "indirect metrics" such as  \gls{flop} or memory footprint can serve as a starting point but do not take into account the platform-specific non-linearities. As a result, networks optimized towards "direct metrics" considerably outperform "indirect" optimized architectures in terms of the selected metrics~\cite{Yang2018a,Cai2019}. On the one hand, the enormous design space for neural network architectures makes it difficult to design a network that runs at high efficiency on all hardware architectures. On the other hand, not all networks work with the same efficiency on a given platform. 
For example, Fig.~\ref{fig:networks} shows the effective compute performance when running 12 networks used for evaluation in this paper on a ZCU102 Xilinx MPSoC evaluation board. Furthermore, the computational roofline shows the maximum reachable, effective compute performance.

We can see the high variance of the effective compute performance for a variety of different network architectures when executed on the same hardware. Due to the large differences in effective compute performance, we can conclude that it is not sufficient to divide the number of operations of a network by the peak compute performance of the target device to achieve a satisfying estimation of the network execution time. So when aiming for field deployment, it is difficult to choose a specific hardware platform before deciding on the network architecture. As a result, there have been some recent attempts to predict network latency and performance on different hardware platforms. However, most of the work targets either \glspl{gpu}~\cite{paleo, neuralpower} server or the embedded \glspl{cpu}~\cite{Yao2018, previous}, leaving out a wide range of hardware accelerators such as \glspl{fpga} and hardware specifically designed for AI tasks e.g. Xilinx ZCU102 and Intel \gls{ncs2}. In this work we aim to model performance of such \gls{dnn} hardware accelerators.
Also, the existing work does not take into account the graph optimizations undertaken by the compiler, which leads to changes in the accuracy of the prediction. 

Therefore, we propose a framework for the generation of stacked, mapping models and layer models to estimate the network execution time. To our knowledge, this is also the first work in which the different approaches to modeling layer execution time and mapping models are systematically investigated and evaluated on a broad range of network architectures.

This paper makes the following key contributions:
\begin{itemize}
    \item We introduce \gls{annette}, a time estimation framework that allows predicting the execution time of Deep Neural Networks on hardware accelerators based on a stacked modeling approach of mapping models and layer-wise estimation models
    \item We propose mixed models for layer execution modeling to decrease the necessary model complexity of the statistical models to cover also computational utilization inefficiencies
    \item We propose a methodology to extract mapping models and layer execution models from micro-kernel and multi-layer benchmarks. Our evaluation of the generated mapping models and layer models on a set of 12 state-of-the-art models show a mean absolute percentage error of 3.41\% for the ZCU102
    \item We compare mixed layer models with statistical layer models, the roofline model, and a refined roofline model in terms of accuracy and fidelity
\end{itemize}




\section{Related Work}
\label{sec:relatedwork}

Several studies have been performed to measure how well certain DNNs perform on different hardware. Their purpose is to explore the design space and to get the highest efficiency out of the hardware. In EmBench \cite{Almeida2019}, common DNNs like ResNet, ShuffleNet, and MobileNet were tested on a wide range of hardware, ranging from power consuming server hardware like the NVIDIA GeForce RTX 2080 Ti GPU to mobile devices like the Intel NCS2. A key finding in EmBench was the Pareto curve of accuracy and latency of different networks on the hardware devices. Often, it depends on the type of layers used in the respective networks. While they tested all different combinations of networks and hardware, our work takes another approach. For each hardware, we provide a method that measures latency for each layer type and then estimates the latency of a whole composed network. Together with known accuracies of the architectures in NASBench \cite{Ying2019}, we can then explore the Pareto curve of a specific hardware platform without further measurements. 

MLPerf \cite{reddi2019mlperf} is an attempt by over 30 organizations to create an industry-wide standard benchmark to assess the vast number of machine learning software and hardware combinations, while DAWNBench \cite{coleman2017dawnbench} is led by academia. MLPerf limits the problem space by defining a set of scenarios, datasets, libraries, frameworks, and metrics. Additionally, it specifies prohibited operations to enhance comparability under equal terms. For our statistical model, MLPerf could provide additional measurements to align it for new hardware and enhance our measurements. However, the available data does not suffice to construct accurate mapping models and layer models.

Besides characterizing accelerator hardware, hardware optimized neural architecture search (NAS) is becoming increasingly popular and powerful. While handcrafted cells of ResNet and Inception lie close to the Pareto optimum at GPUs \cite{Ying2019}, the design space for mobile devices is very large \cite{wu2019fbnet}. It offers potential for automated architecture search, especially when the demand for customized networks rises. FBnetV3~\cite{fbnetv3}, SqueezeNAS~\cite{shaw2019squeezenas} and Proxyless NAS~\cite{Cai2019} focus on low-latency network architecture search for mobile devices. They were developed to replace costly redesign DNNs for certain tasks on certain platforms. While SqueezeNAS focus on semantic segmentation, FBNetV3 and Proxyless NAS focus on classification tasks. Both tools show superior latency-accuracy tradeoffs compared to MobileNet. SqueezeNAS, as well as Proxyless NAS, first generate a super network, in which each cell is selected from a search space. They approximate latencies by building look-up tables for the selected blocks within the design space to save time. All three works could profit from a uniform estimation framework that accurately predicts performance for multiple platforms. In NetAdapt~\cite{icg:yang2018netadapt}, empirical measurements on a Google Pixel 1 CPU are used to construct layer-wise look-up tables to shrink a pre-trained MobileNetV1 until the resource constraints are met to optimize \glspl{dnn} for inference on mobile devices. FBNetV3 uses multi-use predictors to power their neural architecture search algorithm by predicting architecture statistics such as accuracy and the proxy metrics FLOPS and number of parameters.

NeuralPower~\cite{neuralpower} is an attempt to estimate execution latency, power, and as a result, overall energy consumption based on layer-wise sparse polynomial regression for \gls{gpu} platforms. In terms of execution time estimation, NeuralPower achieves an average accuracy of 88.24\% on the networks VGG-16, AlexNet, NIN, Overfeat, CIFAR10-6conv. In addition to the layer-wise time estimation, the same modeling method is also applied to estimate power and finally energy consumption with even higher accuracy.
FastDeepIoT~\cite{Yao2018} uses execution time models based on linear model trees to predict the layer execution time on the devices Nexus 5 and Galaxy Nexus to finally compress VGGNet for both devices and reduce the neural network execution time by 48\% to 78\% and energy consumption by 37\% to 69\% compared with the state-of-the-art compression algorithms.
In PreVIous~\cite{previous}, the execution time models are based on linear regression, and for the devices, Raspberry 3 and Odroid-XU4 reaches about 96\% average accuracy for the layer-wise estimation. These results lead us to believe that the task of estimating layer execution times for task optimized computing architectures is significantly more challenging than for \glspl{cpu}. Therefore, we propose a methodology for generating stacked mapping and layer execution time models for hardware accelerators and systematically compare the prediction accuracy of different modeling approaches.
Other than that, MLPAT~\cite{mlpat} and DNN-Chip Predictor~\cite{chippredictor} propose white box approaches to estimate timing, power and energy. MLPAT reports only 10\% error when predicting the power of the TPU-v1. DNN-Chip Predictor's predicted performance differs from those of measurements of FPGA/ASIC implementation by no more than 17.6\% when evaluated for two \glspl{dnn} on three accelerator architectures.



\section{Annette Architecture}
\label{sec:method}

\begin{figure}[ht]
  \centering
  \includegraphics[width=0.7\textwidth]{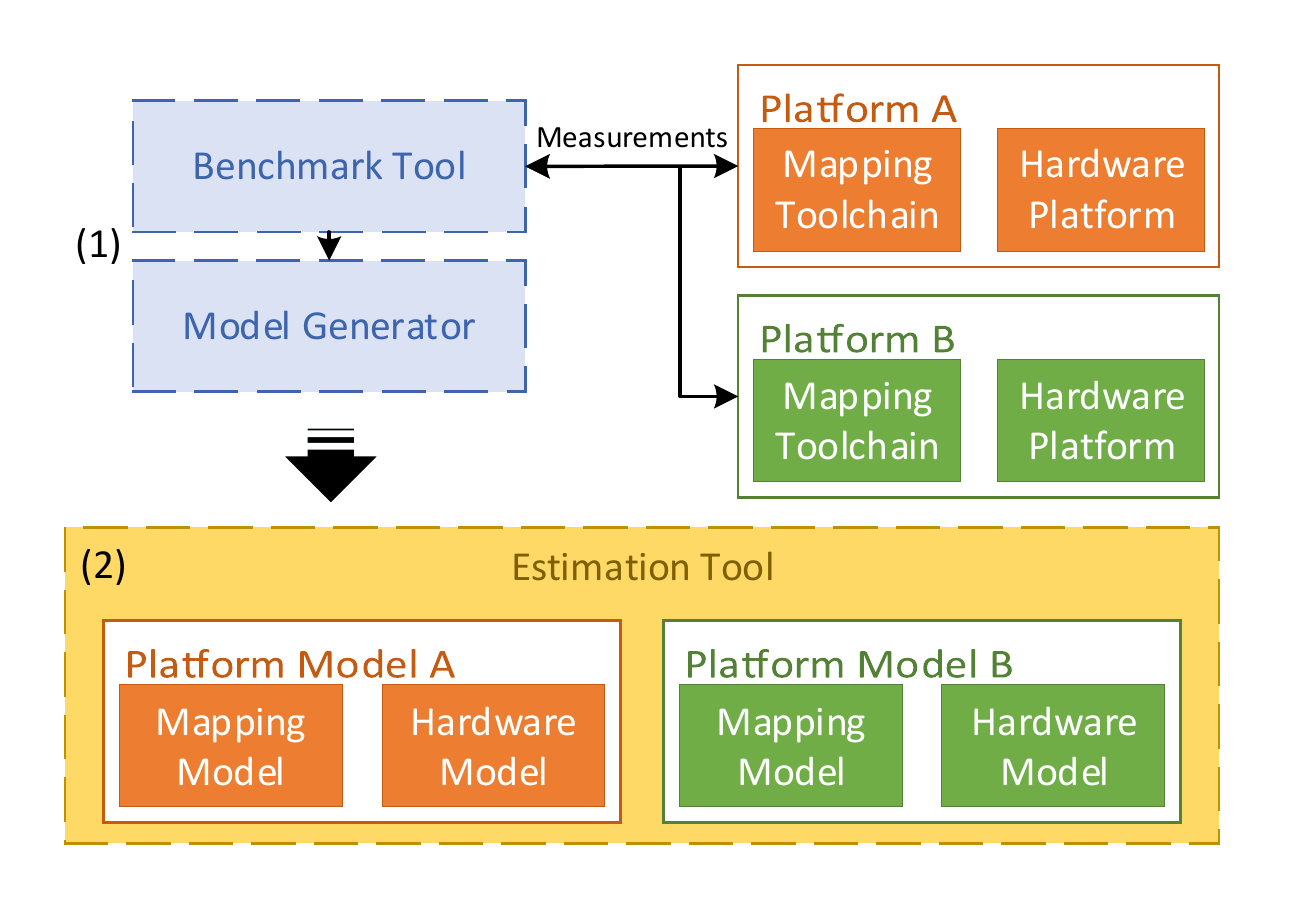}
  \caption{Overview of the Annette architecture: In the benchmark phase (1), first the platform benchmarks are performed and then the platform models are generated. In the estimation phase (2), the Estimation Tool reads a network description graph, and provides an estimated network execution time, a detailed layer-wise execution time prediction table, and a predicted execution graph.}
  \label{fig:Overview}
\end{figure}

Fig.~\ref{fig:Overview} shows an overview of the proposed framework, allowing us to generate abstraction models for the hardware platform and the mapping toolchain. The \textbf{Benchmark Tool} generates networks,  which are then optimized by the provided mapping toolchain for the selected platform. During the benchmark phase, we execute the generated models on the target device and extract detailed layer execution times. 
We rely on the provided platform tools for mapping, inference, and profiling. With the collected profiling information, the \textbf{Model Generator} can create abstraction models of the graph optimizations and the different layer types. These abstraction models are used in our \textbf{Estimation Tool} to predict the performance of a network without compiling and executing the model. Furthermore, detailed insights are gained to produce efficient networks for the modeled hardware devices. 

\section{Benchmark Tool}
\label{sec:benchmark}

\begin{figure}[ht]
  \centering
  \includegraphics[width=0.7\textwidth]{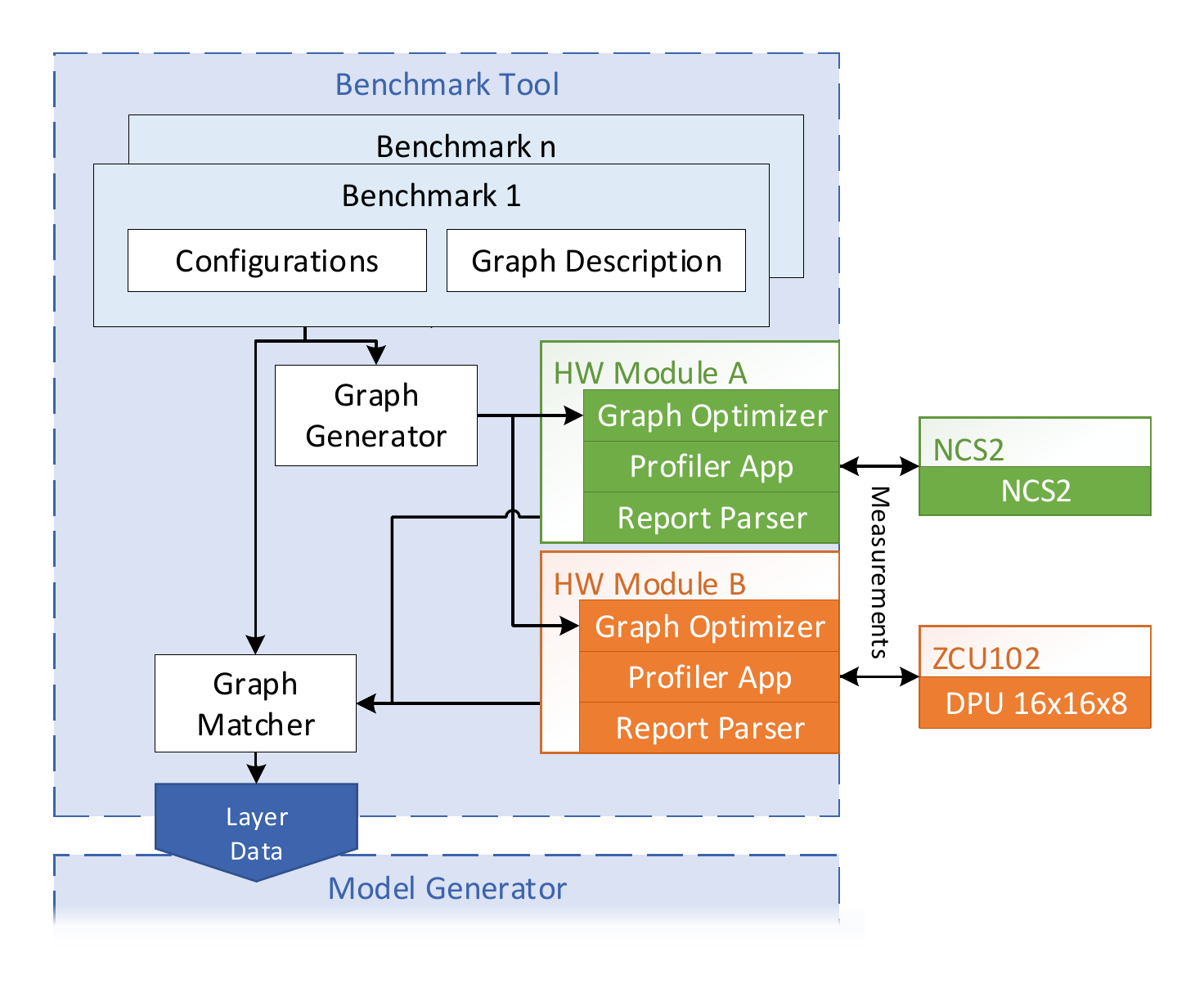}
  \caption{The Benchmark Tool profiles the provided set of benchmark models with different configuration settings on the hardware platforms and generates layer data files}
  \label{fig:benchmark_tool}
\end{figure}

For platform characterization, we make use of two kinds of benchmarks: micro-kernel benchmarks and multi-layer benchmarks. We aim to characterize the computational efficiency of a hardware platform when executing only a specific layer with the micro-kernel benchmarks. On the other hand, multi-layer benchmarks give us a deeper understanding of which kind of layers are executed separately and which layers can be fused, reducing the off-chip data movement.
Fig.~\ref{fig:benchmark_tool} depicts the workflow of the Benchmark Tool.

 We define a benchmark as one parametric network graph profiled several times on the hardware with different input resolutions or kernel-sizes. Each generated network architecture stays the same for each benchmark, while only the layer parameter settings (e.g. number of channels and kernel size) are changed according to the configuration file. The input for each benchmark is a configuration and a graph description file. The configuration file defines the parameter settings of each measurement. The graph description file defines the architecture of the benchmarked dummy network.
The Graph Generator module builds the network models based on the description and configuration information and feeds it to the hardware-specific modules. In each hardware module, the network graph is initially optimized and compiled by the platform mapping toolchain. The optimized graph is then inferred on the target device in a platform specific benchmark application. Then, a report is generated with the help of the platform profiling tool. Finally, the report is parsed into a standard format, and the \textbf{Graph Matcher} compares the collected layer data with the original input network. 

Running each benchmark separately is a time-consuming task of about three to five days per benchmark, as each model must go through the entire compilation toolchain before the desired measurements can be made. Therefore, we have developed some network models that allow us to measure several kernels within a benchmark run. In this case, the models must be constructed in a particular manner so that the compiler cannot fuse layers or that the computational effort for an operation is not increased. It would result in measuring more than just the desired micro-kernel. Those linear network graphs still count as micro-kernel benchmarks since we still measure the execution time of each layer individually.
It must be taken into account that when using graphs with more than one layer for micro-kernel benchmarking, the maximum allowable layer size may be smaller than when measuring a single layer.
The choice of configuration parameters has an additional influence on the benchmark wall time. It also influences the insights that can be gained from the collected data and on the understanding of how to model the accelerator This topic is discussed in Section~\ref{sec:modeling}. 

We use the same mechanism for the multi-layer benchmarks, with the difference that they have more configuration parameters. The goal of these benchmarks is primarily to model the mapping toolchain and understand which optimizations the graph optimization toolchain can perform, but also to be able to benchmark multiple layers at the same time.

\begin{figure}[ht]
  \centering
  \includegraphics[width=0.7\textwidth]{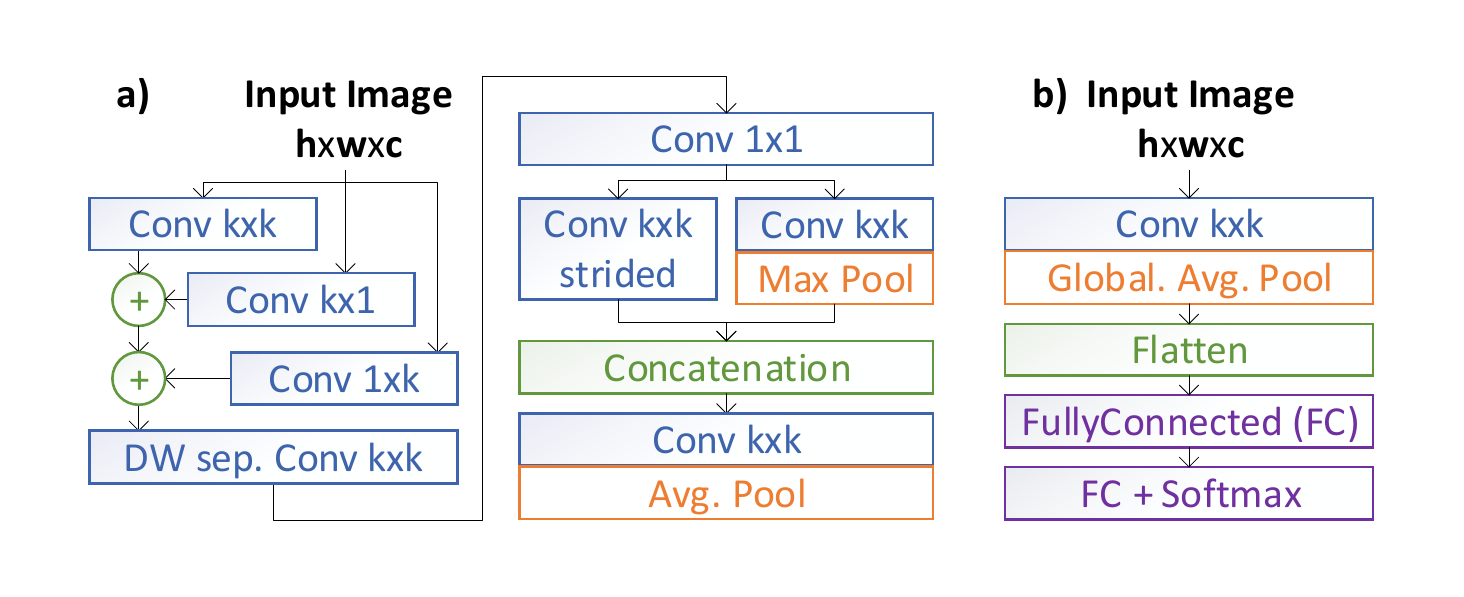}
  \caption{The multi-layer benchmark networks
  (a) ANNETTE ConvNet for characterizing convolution and pooling layers; (b) ANNETTE FCNet for benchmarking global average pooling and fully connected layers}
  \label{fig:benchmark_net}
\end{figure}

The Graph Generator builds the network models, which are benchmarked on the target platform, based on the graph description and the configurations table. It iterates through the configurations table generating one network model per parameter setting.
We apply micro-kernel benchmarks for 2D convolution, 2D depth-wise separable convolution, max pooling, average pooling, and fully connected layers, with values in the range from 8 to 2048 for \gls{height}, \gls{width}, \gls{channels}, \gls{filters}, input and output neurons, kernel sizes ($k_h$, $k_w$) 1, 3, 5, and 7, and pooling sizes from 2 to 10, resulting in a total of about 35k measurements per layer.
Figure \ref{fig:benchmark_net} illustrates the network architectures used for the multi-layer benchmarks. All convolution layers are followed by batch normalization and ReLU layers.

The \textbf{Hardware Modules} are simple scripts that automatically call the platform optimization (Graph Optimizer) and compilation toolchain to prepare the benchmark models for inference. In the case of \gls{dnndk}, as a developing framework for the hardware module \gls{dpu} on the ZCU102 MPSoC board, optimization and compilation functionality are provided through the Deep Compression Tool (DECENT) and the Deep Neural Network Compiler (DNNC) respectively~\cite{dnndk}. In the case of \gls{ncs2} the graph is optimized and compiled by the OpenVINO Toolkit~\cite{openvino2018}. Similarly, we rely on provided execution and the platform specific profiler applications (Profiler App) to extract the layer execution times for the compiled networks. To avoid measurement errors, we average the results of 20 iterations. Finally, a Report Parser extracts the layer-specific information and maps it back to the original graph, comparing the executed layers with the original layers by their names. Therefore, the Profiler App must provide execution times and layer names. The execution information is stored in a standardized format so that the Graph Matcher can process the provided data in the same way for each platform. These encapsulated hardware modules make it easy to add future hardware to the benchmarking tool.

In addition to the Report Parser, the Graph Matcher extracts information about the differences between the original input graph and the final net graph executed on the target device. While the parser merely ensures that there are no changes to the original naming scheme and provides a standardized output, the Graph Matcher extracts additional information about the optimization behavior of the mapping toolchain. The Graph Matcher creates a layer result file for each executed layer and an optimization mapping file for the entire benchmark. The layer result files contain information about the layer parameters, e.g., height, width, number of input channels, and the resulting execution times. To track the behavior of the mapping toolchain, we also store ternary variables that successive layers have been fused with the measured layer. This merging variable can store the following states: not-fused, fused, and possibly-fused. Possibly-fused is used because, it is not possible to detect where the layer has been merged or not for layers with multiple inputs.

\begin{figure}[ht!]
  \centering
  \includegraphics[width=0.6\textwidth]{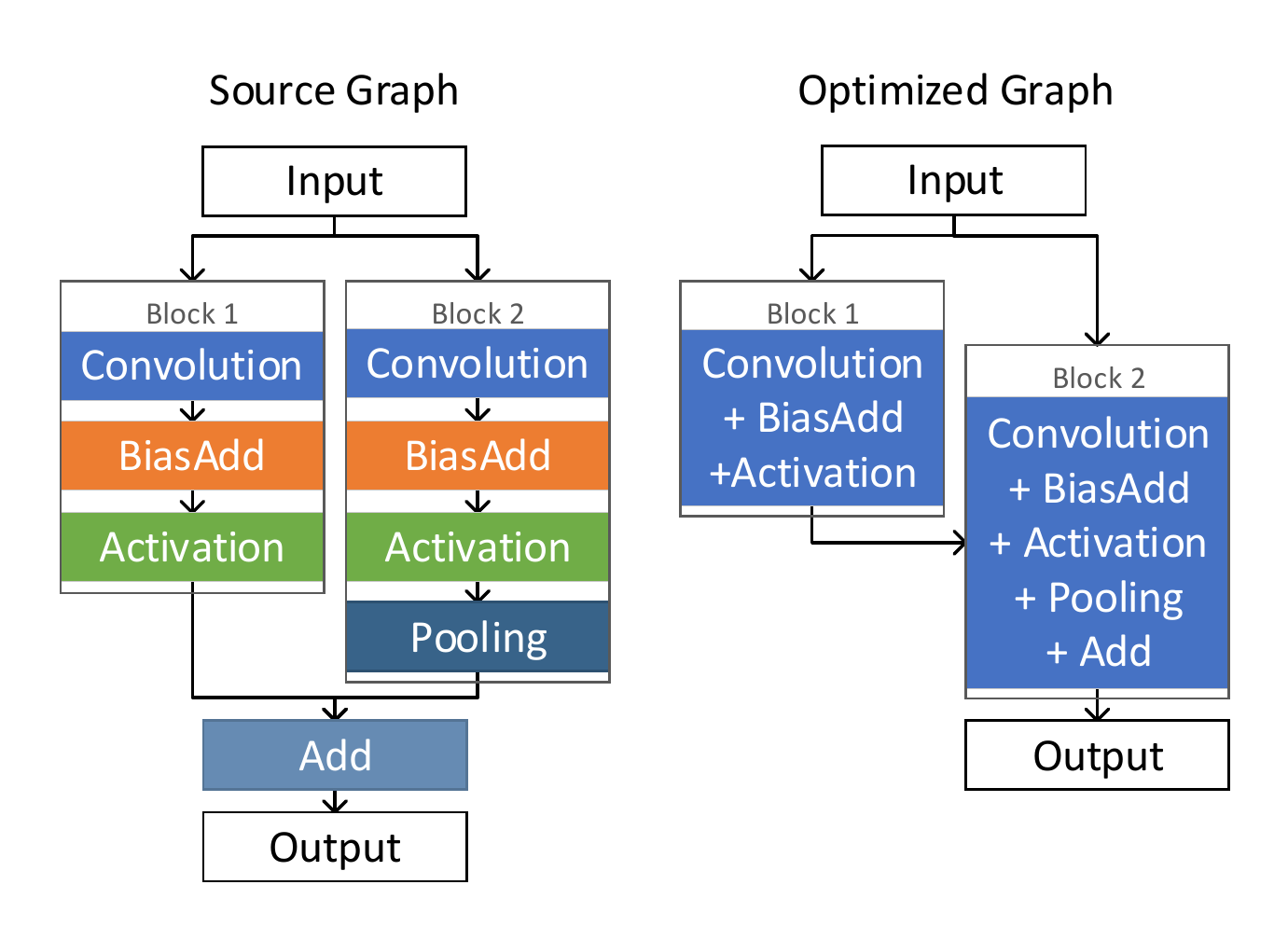}
  \caption{Graph Optimization}
  \label{fig:graphopt}
\end{figure}

Fig.~\ref{fig:graphopt} shows an example of how a graph could be
optimized by the mapping toolchain. In the specific example we set the
fused flags for \textit{BiasAdd, Activation} operation in the
\textit{Convolution} layer of block 1 and 2 to \textit{fused}. In
block 2, the fused flag for the \textit{Pooling} operation is set to
\textit{fused} as well. Here, it is important to note that since the
pooling layer also has a set of parameters, i.e., pooling height,
pooling width, pooling stride, and pooling type that define its
execution policy, we also need to add those parameters to the already
existent stored parameters for the convolution layer. It enables the
graph optimizer modeler to extract rules that define in the case of which
parameter combinations the layers can be fused. Since the element-wise
addition layer may have been matched to either block 1 or block 2, the
fused flag for the \textit{Add} operation is set to
\textit{possibly-fused} in both blocks.

The generated layer data consists of a table for each layer type that for each measurement contains the parameter settings of the layer e.g. height, width, channels, kernel size as well as the measured execution time.
This data is then fed to the Model Generator to extract optimization and layer models for the final estimation step.

\section{Model Generator}
\label{sec:modeling}
This section explains how we model the graph optimizations of the mapping toolchain and the computational efficiency of the hardware platforms to achieve better overall latency estimation accuracy.
As depicted in Fig.~\ref{fig:networks}, not all networks are
computed with the same efficiency when compared to the number of
operations in the convolution layers. There are two leading causes of
the non-linear nature of the relationship between the number of operations
and execution time. First, the non-convolutional layers cannot be neglected.
They are not considered in the commonly claimed number of operations, such as element-wise addition,
concatenation, activation, or pooling. It is crucial for the execution time of these layers, whether they are executed in isolation or connection with a convolution layer \cite{alwani2016fused}. The second factor is that the utilization of computational resources for the same layer can depend on the parameter settings on a specific layer (e.g., height, width). It means that two compute-bound layers with the same number of operations but with differently shaped input and weight tensors are not necessarily computed with the same efficiency~\cite{Yao2018}.
\begin{figure}[ht!]
  \centering
  \includegraphics[width=0.6\textwidth]{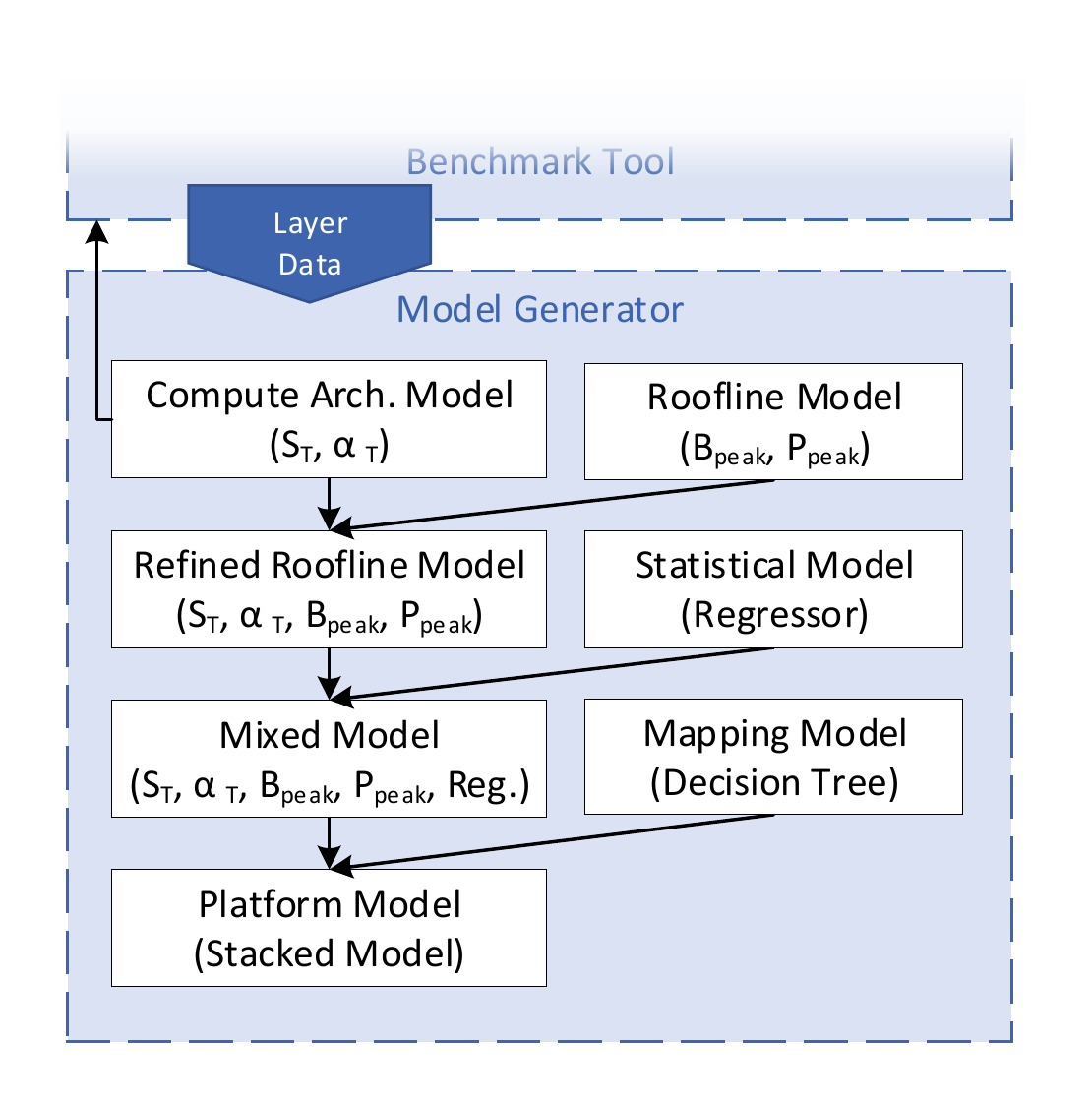}
  \caption{The Model Generator, extracts a stacked model consisting of mapping models and layer models}
\label{fig:generator}
\end{figure}

To cover all these aspects, we propose a stacked model approach to model the overall network execution time accurately. Fig.~\ref{fig:generator} shows how the different models are fused for the generation of the platform model. Tab.~\ref{tab:Glossary} describes the parameters for the models extracted from the benchmarks. The first performed benchmarks are input parameter sweeps to determine the unrolling parameters $\vec{s}$ and $\vec{\alpha}$. These parameters describe the amount of parallel performed multiplications in per dimension of the compute architecture and the parallelization efficiency. With the help of these two parameter vectors, we can construct a model that describes the utilization efficiency of several compute architectures (e.g. systolic arrays). Additionally, preliminary values of $P_{peak}$ and  $B_{peak}$ are determined, which describe the peak performance and the peak off-chip bandwidth. 

These parameters are determined automatically based on measurements or knowledge of the computing architecture. Once determined, the parameters are fed back to the Benchmark Tool to adjust the parameter settings for the succeeding benchmarks. The rest of the micro-kernel benchmark results are used to generate the \textbf{Roofline Model} by deducing the final values of $P_{peak}$ and  $B_{peak}$, which together with the previously determined unrolling parameters, construct the \textbf{Refined Roofline Model}. We combine the \textbf{Statistical Model} and the \textbf{Refined Roofline Model} in the \textbf{Mixed Model}. For the final \textbf{Platform Model}, we add the \textbf{Mapping Model}, which covers optimizations performed on the graph before the actual execution.

\subsection{Layer Execution Time Models}
\label{sub:m_layer}
For the construction of layer-level execution time models, we rely on
the measurements performed in the benchmarks. We construct parametric
analytical models for the \textit{convolution}, the \textit{depth-wise separable convolution}, the \textit{fully connected}, and \textit{pooling layer}. The selection of these layers is motivated, similarly as in the works \cite{paleo,neuralpower}, by the fact that these are the most computational intense layers and, therefore, most critical. However, we will also show that it is also crucial for more complex network architectures to model different layers to achieve accurate results with high fidelity. While the simple roofline model describes most layers with satisfying accuracy, we refine the roofline model for the convolution layer to increase the estimation accuracy.


\begin{table}
\centering
\caption{Model parameters}
\label{tab:Glossary}
\begin{tabular}{l|l}
\hline
         Parameter &        Description  \\
\hline
\hline
 \textbf{HW par.} &  \\
 $P_{peak}$ &     Peak performance of the Architecture (ops/sec)\\
 $B_{peak}$ &     Peak off-chip bandwidth (byte/sec)\\
 $\vec{s}$ &        Spatial unrolling parameter vector of the Architecture\\
 $\vec{\alpha}$ &   Spatial unrolling efficiency coefficient vector\\
\hline
 \textbf{SW par.} &  \\
 $\vec{x}_{n}$ & Input feature vector describing layer $n$ \\ 
 $f_n$ & Number of operations in layer $n$ (ops) \\
 $D_n$ & Data transferred in layer $n$ (bytes) \\ 
\hline
 \textbf{Interm.} &  \\
 $u_{ef\!f_n}$ & Utilization efficiency of comp. resources in layer $n$ \\ 
 $u_{stat_n}$ & Statistically computed efficiency in layer $n$ \\ 
 $P_{ef\!f_n}$ & Effective performance in layer $n$ (ops/sec)\\
\hline
 \textbf{Predictions} &  \\
 $\hat{T}_{n}$ &     Estimated time of layer $n$  (sec)\\
\hline
\end{tabular}
\end{table}

\subsubsection{Analytical Models}
\label{sub:m_l_ana}




For the estimation framework to always work with at least the most simple model, we implement the roofline model \cite{williams2009roofline} for all layer types as a fallback solution. 
In the roofline model for each layer $n$, smallest achievable execution time is either limited by the peak computational performance $P_{peak}$ or the maximal bandwidth $B_{peak}$. In layer $n$, with the data to be transferred $D_n$ and the number of operations $f_n$ give us the estimated execution time $\hat{T}_{roof_n}$ with the effective computation performance $P_{ef\!f}$ equal to $P_{peak}$
\begin{equation}
\label{eq:time_roofline}
\hat{T}_{roof_n}(f_n,D_n) = max(\frac{f_n}{P_{peak}}, \frac{D_n}{B_{peak}}).
\end{equation} 
Keeping in mind that for fused layers the term of $D_n$ has to be corrected (see Section~\ref{sub:m_optimizations}), this formulation of the roofline model can be applied to the four named layer types and will be denoted in the experimental section as roofline model. 

However, as mentioned earlier, computational efficiency also depends on
how the shapes of the input-, weight- and output tensors are mapped on the computing architecture.
When incorporating the reduced utilization efficiency $u_{ef\!f}$ in equation~(\ref{eq:time_roofline}) we obtain
\begin{equation}
\label{eq:refined_roofline}
\hat{T}_{ref_n}(f_n,D_n) = max(\frac{f_n}{P_{peak} u_{ef\!f_n}}, \frac{D_n}{B_{peak}})
\end{equation}
Next we aim to describe the utilization efficiency of a general compute architecture with an array of \glspl{pe}. The number of spatial dimensions $A$ and the number of \glspl{pe} alongside each dimension $\vec{s} \in \mathbb{N}^A$ define the compute architecture. For example an array could be described with $A = 2$ and $\vec{s} = \begin{pmatrix} 16 & 12 \end{pmatrix}$, which amounts to a total of 192 \glspl{pe}. When computing  a layer, the operations have to be mapped onto the array either spatially or temporally. With the parameter settings of the layer as the feature vector $\vec{x}$ we can approximate the utilization efficiency with
\begin{equation}
\label{eq:util_eff}
u_{ef\!f}(\vec{x}) = \prod^{A}_{i=1} \frac{x_{i}/s_i}{\lceil x_i/s_{i}\rceil}.
\end{equation}
Hereby the size of the vector $\vec{x}$ does not have to match the size of the vector $\vec{s}$ as the operations can also be mapped in the temporal dimension. For example, when mapping a 2D 1x1 convolution layer with a 12x6x128 input feature map and 256 output channels, the feature vector describing the layer could be any permutation of $\begin{pmatrix} 12 & 6 & 128 & 256 & 1 & 1\end{pmatrix}$ depending on the mapping of the layer onto the array. With equation \eqref{eq:util_eff}, for the presented example case and the input feature map height and width mapped spatially onto the 16x 12 array, we would get $u_{ef\!f} = 0.375$.

It has to be mentioned that equation~\eqref{eq:util_eff} neglects the overhead of control units and warming up as well as possible input parameter augmentation for $x_i < s_{i}$. For example, since the first layer in most \glspl{dnn} has three input channels ($x_i = 3$), channel augmentation can often improve performance in the first layer of the neural network.
To allow for further adjustment of the model to different efficiencies for each element of $\vec{s}$ we add the unrolling efficiency vector $\vec{\alpha}$ to get the final utilization efficiency of the refined roofline model

\begin{equation}
u_{ef\!f}(\vec{x}) = \prod^{A}_{i=1}(\alpha_i+\frac{\lceil x_i/s_{i}\rceil}{x_{i}/s_i}(1-\alpha_i)) ^{-1}
\end{equation}

\noindent where $\vec{\alpha} \in \mathbb{R}^A \mid 0 \leq \alpha_i \leq 1$.
The coefficients $\alpha_i$ adjust the impact of the spatial unrolling. According to the terminology used in~\cite{eyeriss2}, $\alpha_i$ allows us to adjust the impact of \textit{spatial} and \textit{temporal fragmentation} on the overall utilization efficiency. So far, we have identified no other method to derive the values of $\vec{\alpha}$ from the system architecture than by measurement.

This refined version of the roofline model allows us to model not only the reduced utilization efficiency of n-D convolutions due to the mapping restrictions of existing compute architectures. It can also be used to model jumps in utilization efficiency caused by higher-level features such as the number of input parameters, weights, or outputs.

We apply the simple roofline model with separately measured data throughput rate and peak performance to the pooling, depth-wise separable, and fully connected layers, respectively, under the presumption that accuracy does not have to be as high as for the convolutional layers. However, it is still important to also capture the execution time of those layers. Furthermore, for fused layers, we define the first term of equation~(\ref{eq:refined_roofline}) as the sum of the execution time of the convolution layer and the following fused layer. For the second term, we adjust the number of transferred data to the overall amount of the fused layer. For example, a convolution layer with a succeeding pooling layer with a stride greater than one has a reduced number of output parameters.

Within the modeling framework, we determine model parameters $P_{peak}$, $B_{peak}$, $\vec{s}$ and $\vec{\alpha}$ for all layers automatically based on the measurements of the Benchmark Tool. At first, we perform sweep benchmarks to measure the layer execution time while sweeping each of the parameters describing the layer. For example, in one sweep for a 2D convolution layer, we measure the execution time, incrementing the number of input channels in each measurement. These sweeps are performed for each parameter at multiple points, while the other layer parameters are set to the same value for the entire sweep. Based on these measurements, we can extract the preliminary values of $P_{peak}$ and $B_{peak}$ by finding the maximum performance and data throughput values.  
Next we determinate the values of $s_i$ and $\alpha_i$, by fitting equation \eqref{eq:util_eff} to the collected data using mean square minimization, with the conditions $\vec{\alpha} \in \mathbb{R}^A \mid 0 \leq \alpha_i \leq 1$ and $\vec{s} \in \mathbb{N}^A$. Lastly with the determined values of $\vec{s}$ and $\vec{\alpha}$ we perform the rest of the benchmarks using preferably layer settings with~\eqref{eq:util_eff} to determine the final values of $P_{peak}$ and $B_{peak}$.

\subsubsection{Statistical Models}
\label{sub:m_l_stat}
Apart from the analytical estimation model, we also generate statistical regression models to estimate the performance for all benchmarked layer types. In general, we found that the statistical models produce more precise results when predicting utilization efficiency rather than the resulting execution time. We estimate the utilization efficiency $u_{stat} = f(\vec{x})$ where $ u_{stat} \in \mathbb{R} \mid 0 < u_{stat} \leq 1$ for each layer separately based on a feature vector $\vec{x}$ describing the layer's parameter settings. Similar to \cite{Yao2018, neuralpower} we include higher-level features such as the number of input parameters and the number of operations. For example, for the 2D convolutional layer we select the feature vector $\vec{x} = (h,w,c,f,k_h,k_w,stride,\#ops,\#in,\#out,\#weights)$.

We applied random forest regression for the statistical models of the network layers, which worked best for the data collected in the benchmarks. Although tree-based regression methods generally do not extrapolate well, they have the useful property that the output values do not explode but remain constant when the input values are outside the training data range. In the case of the $u_{stat}$ estimate, this behavior does not degrade the quality of the estimate. For the final prediction of the layer execution time, we then apply the roofline model with statistically computed utilization efficiency:
\begin{equation}
\label{eq:mixed_model}
\hat{T}_{mix_n}(f_n,D_n) = max(\frac{f_n}{P_{peak}u_{stat_n}}, \frac{D_n}{B_{{peak}_n}})
\end{equation}

Due to the large number of architectural parameters for the convolution layer, we have to carefully select for which configuration parameter settings to perform the measurements. This is important since the points of measurement influence the quality of the resulting statistical models.
To find the best points of measurement for our statistical model, we generate three datasets. For the first dataset, we aim to model the surface of points with the best utilization efficiency. Therefore, we reduce the space of measurements to points with utilization efficiency equal to 1. For the generation of the second dataset, we add Gaussian noise to the parameters with $s_i > 1$ to also cover cases with utilization efficiency $< 1$. The third dataset is the union of datasets 1 and 2.

The experimental results show that, depending on the selected statistical model, too large amounts of measurement points would be required to model the entire surface of dataset 3 correctly. Therefore, we use dataset 1 for the generation of the statistical models and follow a third approach. We combine the generated statistical models with the refined roofline model from Section~\ref{sub:m_l_ana} to achieve higher accuracy for the points with utilization efficiency $< 1$.

\subsubsection{Mixed Models}
\label{sub:m_l_mixed}
To combine the advantages of the statistical and analytical models, we also implement a mixed modeling approach by stacking the statistical model and the refined roofline model. The execution time of the mixed model $\hat{T}_{mix}$ for the layer $n$ can be expressed as

\begin{equation}
\label{eq:mixed_model}
\hat{T}_{mix_n}(f_n,D_n) = max(\frac{f_n}{P_{peak} u_{ef\!f_n}u_{stat_n}}, \frac{D_n}{B_{peak}})
\end{equation}

\begin{figure}[ht!]
  \centering
  \includegraphics[width=0.6
  \textwidth]{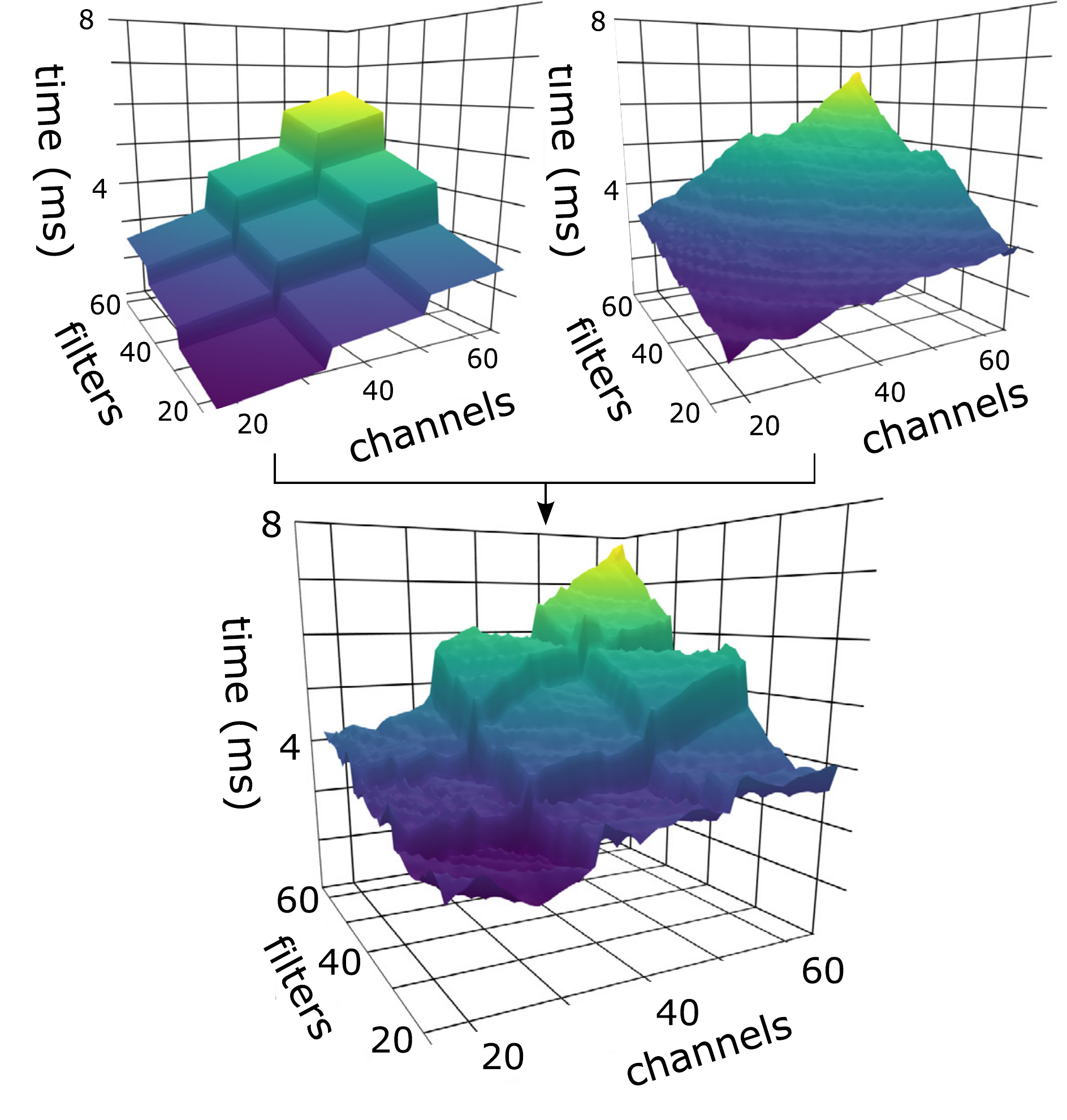}
  \caption{An example of predicted execution time surfaces for the refined roofline model (top left), statistical model (top right) and mixed model (bottom). The plane of the mixed model is an overlay of the refined roofline model and the statistical model.}
  \label{fig:mixed}
\end{figure}

Decoupling the modeling of $u_{ef\!f}$ and $u_{stat}$ has the advantage that the necessary model complexity for the estimation of $u_{stat}$ is reduced, as the model only needs to correctly estimate the points with $u_{ef\!f_n} = 1$. Fig.~\ref{fig:mixed} shows how combining the statistical model and the refined roofline model results in the mixed model.

The analytical part of the model, namely the refined roofline model, covers the step-wise linear shape of the target surface. Based on the refined roofline model, we can determine at which points we want to perform measurements for the statistical model. As mentioned in section~\ref{sub:m_l_stat}, we only select points with $u_{ef\!f} = 1$ for computing the regression model for $u_{stat}$. Therefore, the refined roofline model improves the statistical model twofold: by refining the area with $u_{ef\!f} \neq 1$ and regarding the selection of points for the measurements.

Due to the better choice of data points, the statistical model will produce a better result with a lower risk of overfitting. This also explains why the regression model based on dataset 1 is outperforming the models with additional data points. However, thanks to the analytical part of the mixed model, we can still model the local shape of the surface. We can say that while the analytical part is responsible for modeling inefficiencies of the computational architecture, the statistical model covers the memory architecture.

\subsection{Mapping Models}
\label{sub:m_optimizations}

\begin{figure}[ht]
  \centering
  \includegraphics[width=0.6\textwidth]{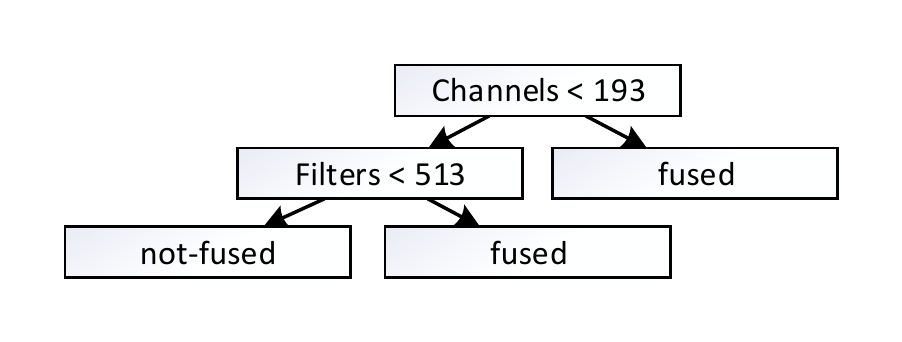}
  \caption{Sample Decision Tree for fusing Pooling and Convolution on NCS2}
  \label{fig:opt_tree}
\end{figure}

The last estimation module we present is the mapping model. The main objective is to predict whether two successive layers have been fused or not. This is important for cases where $T_{total} \ne T_1 + T_2$, where $T_{total}$ is the total execution time of layers 1 and 2; $T_1$ and $T_2$ are the execution times of the two layers when executed separately. As mentioned above, this difference is mainly due to reduced off-chip data transfer and pipelining effects.
For the generation of the mapping models, we use the input feature vectors $\vec{x}$ previously defined for the statistical model and aim to predict the values of the \textit{fused flags} extracted by the Graph Matcher in Section~\ref{sec:benchmark}). We rely on \textit{Decision Tree Classifiers} to determine the rules for the mapping prediction. For example, Fig.~\ref{fig:opt_tree} shows a simplified version of the decision tree for the fusion of a convolution layer followed by a max-pooling layer. We can see that in the example shown, the decision if the two layers are merged or not depends mainly on whether a certain number of \textit{channels} and \textit{filters} in the convolution layer is exceeded or not. We apply the same concept to all fused layer combinations we were able to find in our evaluation networks in  Tab.~\ref{tab:networks}.

\section{Estimation}
\label{sec:estimation}
For the network level estimation, we apply the stacked model presented in Section~\ref{sec:method} on a network description graph. At first, we apply the mapping models to reconstruct the mapping of the platform mapping toolchain. For this, we iterate through all directly connected layers and check whether they should be fused or not. Afterwards, we apply the layer level models on each remaining layer of the optimized graph. The network execution time estimation $\hat{T}_{total}$ is the sum of all estimated layers $\hat{T}_n$.

Because of the different models available for each layer, we implement the estimation framework in a way that we can select the preferred model type but always use the roofline model as a fallback solution so that the highest possible number of layers execution times is always estimated.



\section{Results and Performance Analysis}
\label{sec:results}
To quantify the accuracy of the latency estimation methods presented in Section~\ref{sec:method},  we compare the estimated results to measured times for 12 state-of-the-art \glspl{dnn} listed in Tab.~\ref{tab:networks} from Xilinx Model Zoo~\cite{xilinxmodelzoo} and a randomly selected subset of 34 networks from the models generated in NASBench~\cite{Ying2019} on target devices.

\begin{table}[ht!]
\centering
\caption{Networks used to evaluate estimation accuracy}
\label{tab:networks}
\begin{tabular}{l|l|r}
\hline
         Network &        Dataset &       Operations \\
\hline
\hline
 InceptionV1 &    Imagenet &   3.2G \\
 InceptionV2 &    Imagenet &    4.0G \\
 InceptionV3 &    Imagenet &   11.4G \\
 InceptionV4 &    Imagenet &   24.5G \\
    Resnet18 &    Imagenet &   3.7G \\
    Resnet50 &    Imagenet &    7.7G \\
         Feature Pyramid Network (FPN) &  Cityscapes &    8.9G \\
    Openpose &  AIChallenger &    189.7G \\
 MobilenetV1 &    Imagenet &   1.1G \\
 MobilenetV2 &    Imagenet &   1.2G \\
      YoloV2 &         VOC &     34.0G \\
      YoloV3 &         VOC &  65.4G \\
\hline
\end{tabular}
\end{table}

\subsection{Experimental Setup}
\label{sub:r_setup}
All experiments were performed with batch size 1 to achieve the lowest possible latency, but by adding the batch-size as an additional input parameter for the benchmark dataset and by adding the batch size to the input feature vector of the estimation models, it would also be possible to extend the method to larger batch sizes. For Xilinx \gls{dpu}, we used a ZCU102 evaluation board with a \gls{dpu} configuration of 4096 \gls{mac} units. Measurements on the \gls{ncs2} were performed with an Intel i5-4590 3.3 GHz host processor equipped with 16 GB of RAM in synchronous mode. For both platforms, we used the provided tools for mapping and compilation. To assess the estimator performance, we use two test sets. \textbf{Test set 1} contains the 12 \glspl{dnn} listed in Table~\ref{tab:networks}, and we use it to evaluate in detail the performance for commonly used networks. With \textbf{Test set 2}, we aim to understand whether \gls{annette} could be used for a hardware-oriented neural architecture search. Therefore, we randomly select 34 models of the NASBench~\cite{ying2019bench} neural architecture search dataset, which contains a large variety of different architectures with similar sizes, and evaluate the accuracy and fidelity of our estimator.


\begin{figure}[ht!]
  \centering
  \includegraphics[width=0.65\textwidth]{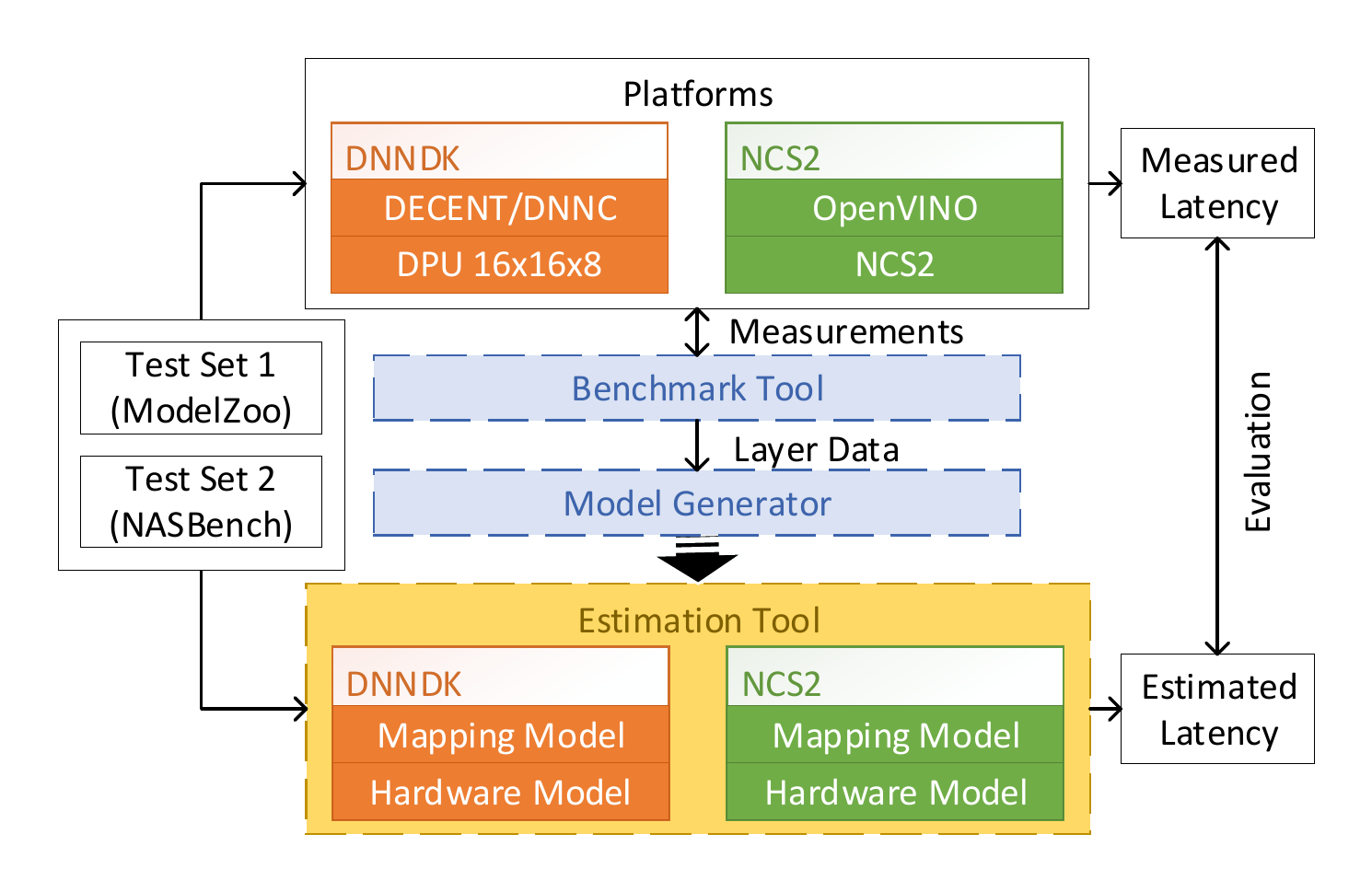}
  \caption{The Experimental Setup for prediction accuracy evaluation}
\label{fig:setup}

\end{figure}
Figure~\ref{fig:setup} shows the experimental setup. In the first phase, the benchmarks from Section~\ref{sec:benchmark} are executed on the target platforms. The execution times of the layers are extracted using the provided profiling tools and stored together with the configuration files of the benchmarks. With the Model Generator (Section~\ref{sec:modeling}), the mapping and hardware abstraction models are derived and made available to the Estimation Tool (Section~\ref{sec:estimation}). In the second phase, the network graphs are fed into the estimator. For evaluation, the resulting estimated times are compared with the execution times measured on the target device. The detailed information provided by the profiling tools allows us to compare not only the total execution times of the networks but also the execution time of each layer.

\subsection{Layer Execution Time Models}
First, we evaluate the accuracy of the previously presented layer execution time models. Tab.~\ref{tab:layer_error} reports the \gls{mae}, the \gls{mape} and the \gls{rmspe} of the different layer models for all convolution layers of the networks in Table~\ref{tab:networks}. The results were estimated and measured for both the \gls{ncs2} and the ZCU102 SoC-board. Additionally, we also report the accuracy of other state-of-the-art execution time prediction methods \cite{paleo,neuralpower}.

\begin{table}[ht!]
\centering
\caption{Layer Execution Time Model evaluation for all convolution layers of the networks in Table~\ref{tab:networks} 
}
\label{tab:layer_error}
\begin{tabular}{rrr|rrr}
\hline
{Work}&Device&{Model Type}&MAE(ms)&RMSPE&MAPE\\
\hline
\hline
\cite{paleo}&Titan X&Analytical&-&{58.29\%}&-\\
\hline

 \cite{neuralpower}&Titan X&Statistical&-&{39.97\%}&-\\
\hline

\multirow{4}{*}{This}&\multirow{4}{*}{NCS2}&Roofline&0.783&63.64\%&32.58\%\\
&&Ref. Roof.&0.730&61.42\%&31.69\%\\
&&Statistical&0.402&44.13\%&15.59\%\\
&&\textbf{Mixed}&\textbf{0.360}&\textbf{42.60\%}&\textbf{15.57\%}\\
\hline
\multirow{4}{*}{This}&\multirow{4}{*}{ZCU102}&Roofline&0.100&18.52\%&39.67\%\\
&&Ref. Roof.&0.066&15.45\%&34.57\%\\
&&Statistical&0.064&12.95\%&16.69\%\\
&&\textbf{Mixed}&\textbf{0.036}&\textbf{10.55\%}&\textbf{12.71\%}\\
\hline
\end{tabular}
\end{table}

The mixed model outperforms the other model types for both platforms in terms of \gls{mae}, \gls{mape} and \gls{rmspe}. 
It is noticeable that for the ZCU102, the refined roofline model has a lower \gls{mae} than the statistical model. Since the \gls{mae} is a non-weighted error metric, we conclude that for the ZCU102, the refined roofline model predicts larger layers more accurately than the statistical model.

For fair comparison to other state-of-the-art works, it has to be mentioned that the reported numbers were measured on a different set of networks\footnote{Paleo and Neuralpower on VGG-16, AlexNet, NIN, Overfeat, CIFAR10-6conv} and for a different set of target devices. While the Paleo~\cite{paleo} and NeuralPower~\cite{neuralpower} target server GPUs (Titan X), our work targets prediction for specific accelerators for neural networks.
However, even in this case, the statistical prediction method outperforms the analytical model. Nevertheless, analytical models are easier to understand and can be easily adapted to similar architectures, whereas a statistical model can only be based on measurements.
Additionally, we applied the NeuralPower estimation method with our collected data for the NCS2 and ZCU102, but we were not able to produce any useful results with a \gls{mape} lower than 1000\%, so we don't list the results of this approach in Tab.~\ref{tab:layer_error}. To our mind, these results are a consequence of the bad extrapolation behavior of polynomial functions, which are used for estimation in NeuralPower.

\subsection{Mapping Models}

We evaluate the performance of the mapping models on the dataset consisting of the layers from the example networks generated by the Benchmark Tool. For the training data set, we consider only the layer pairs that contain the target layer, e.g., for training the decision tree that predicts whether a pooling layer is fused or not, we include only layer pairs in the data set, at least one of which is a pooling layer. Then we select 80\% of the samples for training and 20\% for validation. Tab.~\ref{tab:fused_error} shows the F$_1$ score and the \gls{mcc} for the fusing of element-wise addition and pooling layers.

\begin{table}[ht!]
\centering
\caption{Mapping Model evaluation for fusing pooling and element-wise addition with a preceding convolution layer}
\label{tab:fused_error}
\begin{tabular}{rr|rrr}
\hline
 Device & Layer Type & Total Samples  &   F1 Score &  MCC   \\
\hline
\hline
\multirow{2}{*}{ZCU102}& Pooling & 31733 & 0.973 & 0.871 \\
& ElemwiseAdd & 6079 & 0.990 & 0.923 \\
\hline
\multirow{2}{*}{NCS2} & Pooling & 14628 & 0.824 & 0.831 \\
&ElemwiseAdd & 21942 & 0.792 & 0.733 \\
\hline

\end{tabular}
\end{table}
Since the F$_1$ score ignores true negatives, the \gls{mcc}, which depends on all four confusion matrix categories, should be preferred for the evaluation of the binary classification \cite{Chicco_2020}. It can be seen that the mapping prediction works quite well for both platforms. However, the prediction for the \gls{dnndk} (ZCU102) for both layer types achieves a higher F$_1$ score and MCC than the prediction for the \gls{ncs2}. We assume that the reasons for this are that the \gls{dnndk} is generally more capable of merging several layers and that the optimization behavior of the OpenVINO toolkit depends more on the architecture of the whole network than only on the parameter settings of the individual layers.

\subsection{Evaluation for Test Set 1}

For evaluation of the generated platform models of the \gls{ncs2} and \gls{dnndk}, we perform the mapping and layer-wise estimation for the models listed in Table~\ref{tab:networks}. Then, we compare the predicted network execution time with the measured time. Table~\ref{tab:network_error} shows the \gls{mae} and \gls{mape} of all presented models for the executed networks for the ZCU102 and \gls{ncs2}.

\begin{figure}[ht!]
  \centering
  \includegraphics[clip, trim=0cm 0.9cm 0cm 0cm,width=0.7\textwidth]{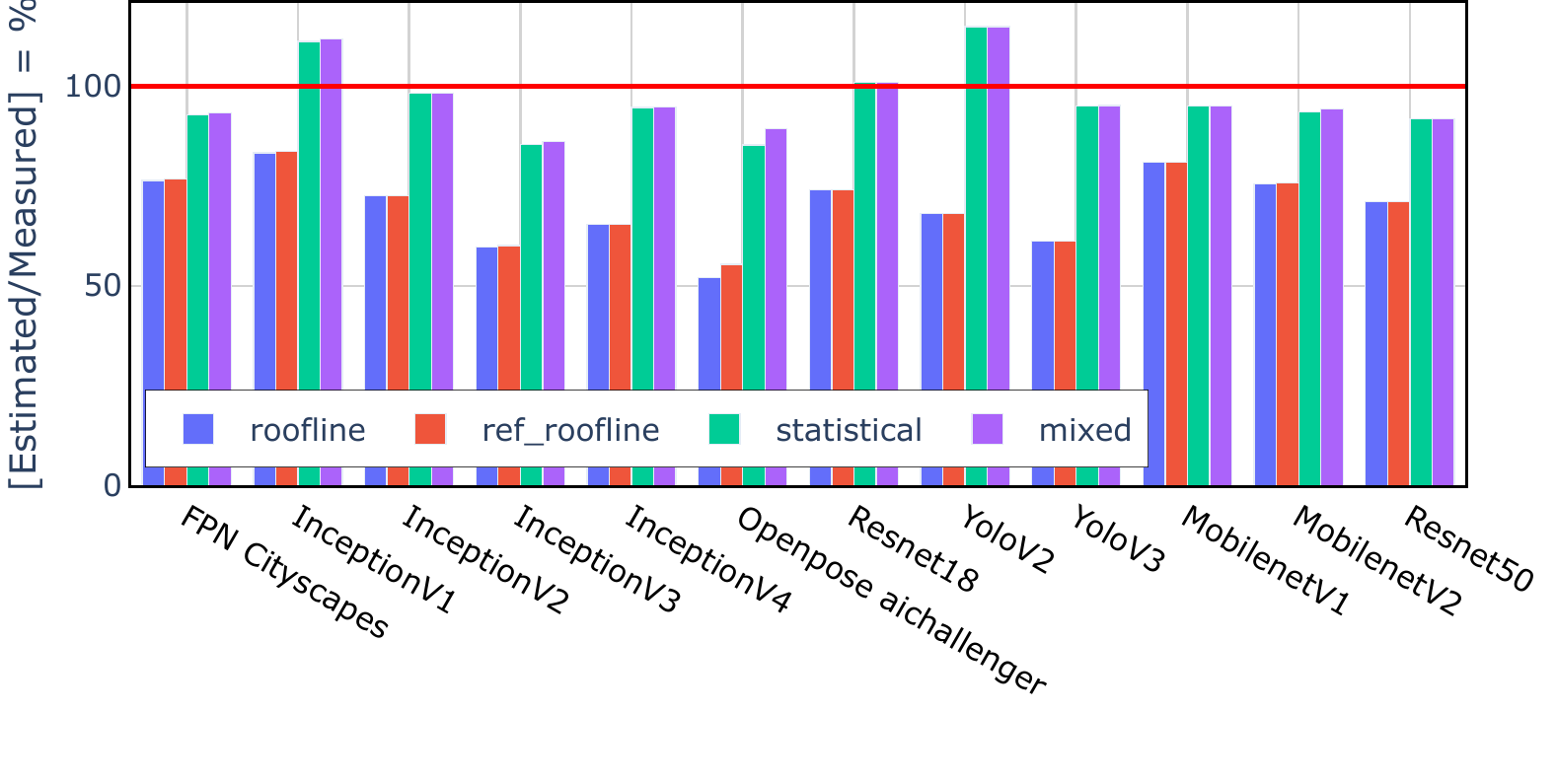}
  \caption{Accuracy of the estimated latency for the selected of Table~\ref{tab:networks} networks on \gls{ncs2}}
  \label{fig:ncs2_lat_xil}
\end{figure}

\begin{figure}[ht!]
  \centering
  \includegraphics[clip, trim=0cm 0.9cm 0cm 0cm, width=0.7\textwidth]{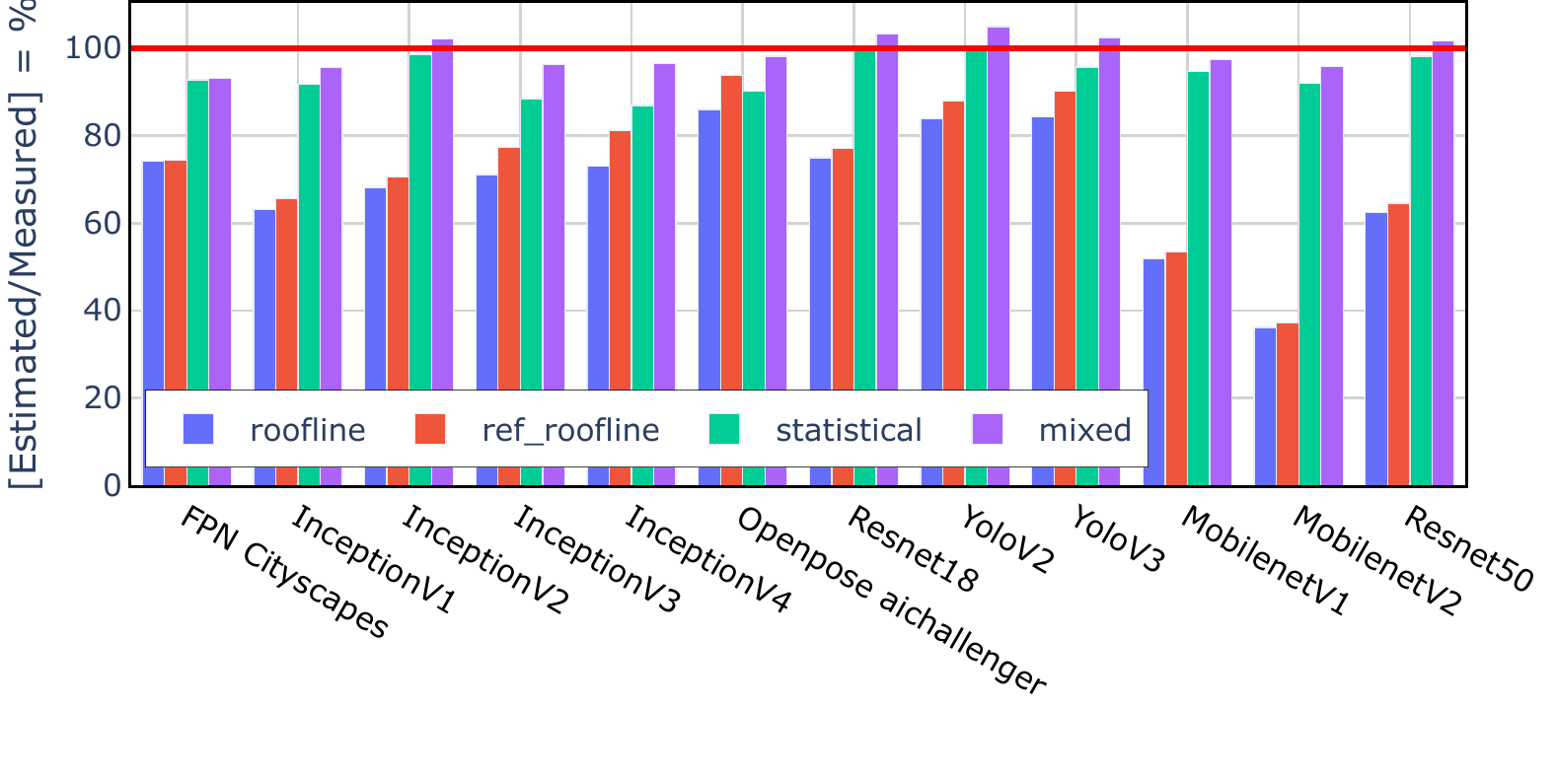}
  \caption{Accuracy of the estimated latency for the selected Table~\ref{tab:networks} on \gls{dnndk}}
  \label{fig:dnndk_lat_xil}
\end{figure}

Fig.~\ref{fig:ncs2_lat_xil} and Fig.~\ref{fig:dnndk_lat_xil} show the estimation accuracy of the platform models. Due to moderate parallelization effects on the \gls{ncs2}, the roofline model and the refined roofline model have similar performance. However, in some cases, the refined roofline model provides slightly better predictions. Also for the \gls{ncs2}, the statistical and the mixed model achieve almost almost similar performance with a \gls{mape} of 7.92\% and 7.44\%, respectively. Overall, the mixed model consistently performs the best for the \gls{ncs2}. Similarly, for the ZCU102, the mixed model provides the most accurate predictions with a \gls{mape} of only 3.47\%. Interestingly, in the case of the ZCU102, for some of the networks, the refined roofline model estimates the network execution time more accurately than the statistical model. Since the refined roofline model mainly covers reduced utilization efficiency due to the computational architecture, we can conclude that for those cases, the main inefficiency lies in the low utilization efficiency of computational resources due to a parameter not aligning with the number of available multiplier resources (see Seciton~\ref{sub:m_l_ana}). The comparison to other state-of-the-art execution time estimators, which are also denoted in Tab.~\ref{tab:networks}, is difficult since the necessary complexity of the model and the resulting accuracy highly depends on the target device. In addition, the evaluation performed in this work includes more complex and larger networks with several different layer types than in other works.

\begin{table}[ht!]
\centering
\caption{Network execution time estimation evaluation for all the networks in Tab.~\ref{tab:networks}. The mixed model outperforms the other models for both platforms in \gls{mae} and \gls{mape}}
\label{tab:network_error}
\begin{tabular}{rrr|ccc}
\hline
Work & Device & {Model Type} & MAE (ms) & MAPE   \\
\hline
\hline
\cite{paleo}& Titan X GPU & Analytical & 23.13 & 27.61\% \\
\hline
\cite{neuralpower}& Titan X GPU & Statistical &    5.11  & 7.96\% \\
\hline
\multirow{4}{*}{This} & \multirow{4}{*}{NCS2} & Roofline 
 &	67.79 & 29.95\% \\
& & Ref. Roofline & 64.93 & 29.56\%  \\
& &  Statistical & 18.52 & 7.92\% \\
& & \textbf{Mixed} &	\textbf{14.95} & \textbf{7.44\%} \\
\hline
\multirow{4}{*}{This} & \multirow{4}{*}{ZCU102} & Roofline & 6.12 & 30.89\% \\
& &  Ref. Roofline & 4.08 & 27.24\% \\
& &  Statistical & 2.49 & 5.97\% \\
& &  \textbf{Mixed}  & \textbf{0.87} & \textbf{3.47\%} \\
\hline

\end{tabular}
\end{table}

\subsection{Evaluation for Test Set 2}
\label{sub:r_nas}
To evaluate the accuracy of the estimations for design space exploration, we perform the estimation for a randomly selected subset of 34 network architectures generated for the NASBench dataset. We select this dataset since it contains several networks with similar sizes that were constructed for the same task. Therefore it is more appropriate to evaluate the fidelity of the estimation tool on Test Set 2. We assess the performance on Test Set 2 for the \gls{ncs2}, which was performing worse on Test Set 1. Table~\ref{tab:nas_fidelity} provides the \gls{mae}, \gls{mape} and Spearman’s rank correlation coefficient $\rho$ as fidelity metric. A perfect Spearman correlation of +1 occurs when the variables are a perfect monotonically increasing function of each other. This property makes $\rho$ a valid measure for fidelity~\cite{fidelity}.

\begin{table}[ht!]
\centering
\caption{Fidelity and Accuracy metrics for Test Set 2}
\label{tab:nas_fidelity}

\begin{tabular}{l|r|r|r} 
\hline
{} &  Spearman's $\rho$ & MAE (ms) & MAPE \\
\hline
Roofline / Ref. Roofline    &      0.971 &      3.50 & 33.38\%\\
Statistical / Mixed            &      0.988 &        0.53 & 9.65\%\\
\hline
\end{tabular}
\end{table}

\begin{figure}[ht!]
  \centering
  \includegraphics[width=0.75\textwidth]{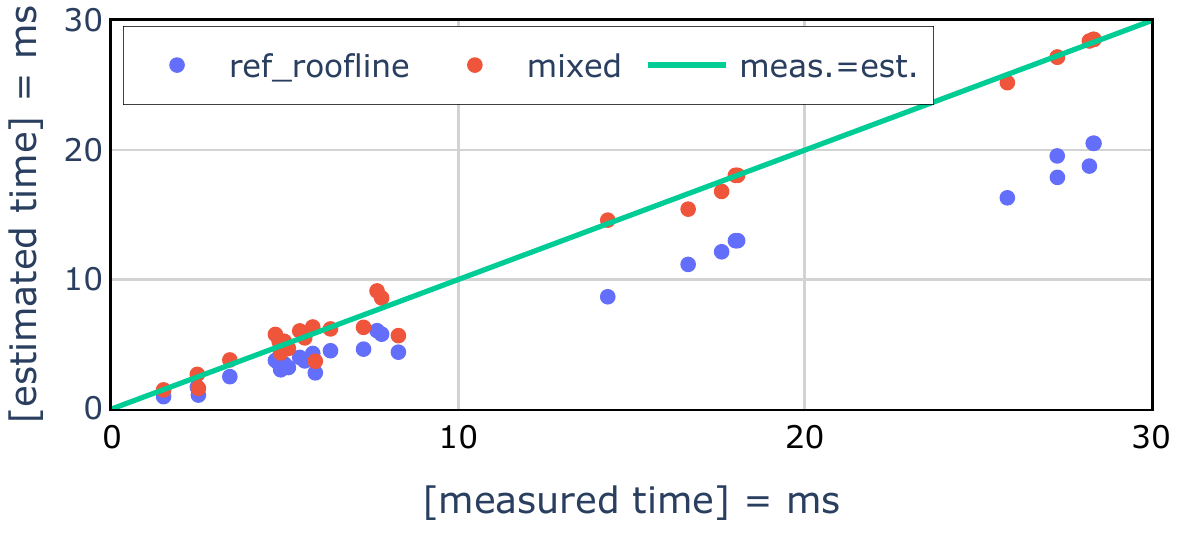}
  \caption{\gls{ncs2} estimation performance for Test Set 2}
  \label{fig:ncs2_lat_nas}

\end{figure}

Fig.~\ref{fig:ncs2_lat_nas} shows the resulting estimated and measured time in milliseconds for the \gls{ncs2}. 
Due to the selected resolutions, there is no difference between the results of the roofline and the refined roofline model. Hence, also the statistical and mixed models achieve the same results. For Test Set 2, the mixed/statistical modeling approach reaches almost a Spearman's rank correlation coefficient of +1 and outperforms the analytic models by more than 20 percentage points in \gls{mape}.

\section{Conclusion}

We propose a framework for execution time estimation for neural network hardware accelerators. It is based on stacked models, consisting of mapping models and mixed layer models. We generate the models based on micro-kernel and multi-layer benchmark results and evaluate the performance on two sets of networks for two selected hardware accelerators. Overall, the mixed models perform best. For a set of 12 state-of-the-art \glspl{dnn}, the estimation with mapping models and mixed models reach a \gls{mape} of only 3.47\% on the Xilinx ZCU102 SoC and 7.44\% on the Intel NCS2 when estimating total network execution times. For the use case of design space exploration, we evaluate the fidelity of the generated models by applying the estimation method on a randomly selected subset of 34 models of the NASBench dataset. The estimation with mapping models and mixed layer models reaches fidelity of 0.988 in Spearman’s $\rho$ rank correlation coefficient metric. The evaluation demonstrates the advantages of applying mixed models for the selected hardware platforms. In the future, we aim to extend the evaluation to additional embedded hardware, such as the Nvidia Jetson platform, to gain additional insights for a different class of accelerators.

Due to the large parameter space of \glspl{dnn}, one crucial point for the development of the estimation framework is to make assumptions about the computing architecture to exclude as many non-meaningful measurement points as possible. An essential clue is the step-wise linear nature of architecture resources, such as an array of multipliers or caches. They follow a linear performance trend until the cache or the multiplier array is fully allocated.
Besides, for a precise estimation, it is important to consider not only the individual layers in isolation but also how they are executed in the overall context. 

We are confident that accurate estimation methods can significantly facilitate informed making of decisions. Nevertheless, it is in the area of neural architecture search where estimation can make a critical contribution to a hardware-specific search or the right choice of networks and hardware in advance of the development of applications.








\bibliographystyle{unsrtnat}
\bibliography{bib/abbreviation-long,bib/references,bib/references-icg}







\end{document}